\documentclass[10pt,twocolumn,letterpaper]{article}

\usepackage{cvpr}              
\definecolor{cvprblue}{rgb}{0.21,0.49,0.74}
\usepackage[pagebackref,breaklinks,colorlinks,allcolors=cvprblue]{hyperref}

\def\paperID{*****} 
\def\confName{CVPR}
\def\confYear{2026}

\title{\vspace{-0.4in}Dual-Route Top-K Retrieval with 1v1 VLM Reranking for the CoVR-R Challenge}

\author{
Yuyang Sun\textsuperscript{1},
Yongliang Wu\textsuperscript{1},
Xingyu Zhu\textsuperscript{2},
Yuxia Chen\textsuperscript{3}, \\
Zhenxiang Jiang\textsuperscript{4},
Yangguang Ji\textsuperscript{4},
Wenbo Zhu\textsuperscript{4},
Yanxi Shi\textsuperscript{4},
Jay Wu\textsuperscript{4},
Shuo Wang\textsuperscript{5},
Xu Yang\textsuperscript{1} \\
\textsuperscript{1}Southeast University
\textsuperscript{2}National University of Singapore
\textsuperscript{3}Independent Researcher \\
\textsuperscript{4}Opus AI Research
\textsuperscript{5}University of Science and Technology of China‌
}

\begin{document}
\maketitle
\begin{abstract}
We describe \emph{Dual-Route Top-K Retrieval with 1v1 VLM Reranking} for the CoVR-R challenge. The method treats composed video retrieval as two coupled problems: finding a sufficiently complete top-k candidate set, and then safely deciding whether any candidate should replace a strong current top-1. We first improve the reasoning/text seed with a VLM slot selector over existing candidates, without introducing DFN visual retrieval. We then add a visual route from contact-sheet embeddings using DFN-H/DFN-L. The routes are merged into a top-10 candidate set, after which a VLM final reranker performs conservative 1v1 comparisons between the current top-1 and each challenger. On the hidden test split, the final system reaches 95.28 R@1, 97.47 R@5, 98.48 R@10, and 99.66 R@50. The main lesson is that CoVR-R benefits more from recall-selection decoupling than from broad text reranking or direct multi-candidate VLM classification.
\end{abstract}

\section{Introduction}

Composed video retrieval (CoVR) asks a model to retrieve a target video given a reference video and a textual modification. CoVR-R makes this setting reasoning-aware: many edits require inferring after-effects such as object state transitions, temporal phase changes, camera/framing changes, or forbidden visual content that should disappear after the edit~\cite{thawakar2026covr}. This makes simple caption overlap or direct text-video similarity brittle.

Our solution is built around two observations from the challenge split. First, the original reasoning/text ranking often contains a strong top-1 but misses visually plausible edited targets in the top ranks. Second, visual retrieval can recover those missing targets, but directly trusting its winner is unsafe. The system therefore needs a candidate-construction mechanism that is aggressive about recall and a final decision mechanism that is conservative about top-1 replacement.

We propose a zero-shot staged pipeline. A reasoning/text route supplies a stable seed ranking, then a VLM slot selector improves top-1 within the existing candidate list before any DFN visual retrieval is introduced. A visual route then retrieves additional candidates from contact-sheet embeddings. The merged top-10 set is passed to a VLM final reranker that performs 1v1 comparisons against the current top-1. This design avoids asking the VLM to solve a broad top-k classification problem and instead asks a narrower question: whether a specific challenger is clearly better than the current best prediction.

Our main contributions are: (1) a staged top-1 selection and dual-route top-k construction strategy for CoVR-R; (2) a 1v1 VLM final reranker with fatal-error and consensus checks for risk-controlled top-1 promotion; and (3) an ablation-backed analysis showing that candidate recall and top-1 selection should be optimized separately.

\section{Method}

\subsection{Overview}

\begin{figure*}[t]
\centering
\includegraphics[width=0.98\linewidth]{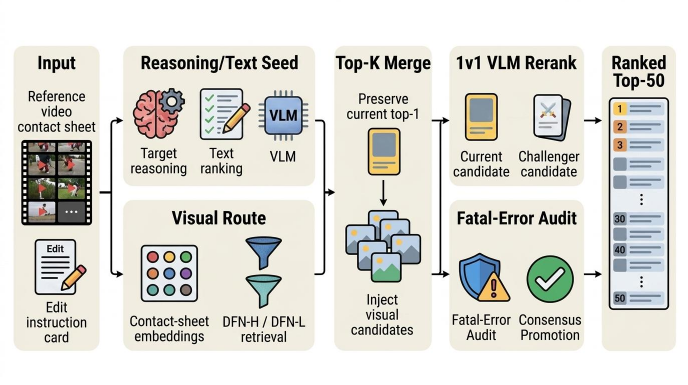}
\caption{Staged top-1 selection and dual-route top-k retrieval with 1v1 VLM reranking. The reasoning/text route is first improved by a VLM slot selector without DFN visual retrieval. The visual route then injects contact-sheet retrieval results into the top-10 candidate set. The final reranker compares the current top-1 against challengers one at a time, using fatal-error and consensus checks before promotion.}
\label{fig:method}
\end{figure*}

Figure~\ref{fig:method} shows the final pipeline. The method first builds a reasoning/text seed, strengthens its top-1 with a VLM slot selector over existing candidates, then constructs a compact top-10 candidate set with a complementary visual route. A 1v1 VLM final reranker decides whether any merged candidate should replace the current top-1. The split is deliberate: candidate construction can be recall-oriented, while top-1 replacement is accepted only under conservative evidence.

\subsection{Route A: Reasoning/Text Seed}

The first route follows a reason-then-retrieve structure. A VLM is prompted with the reference video and edit instruction to infer the target-side state, including required objects, action/state changes, scene constraints, and source details that should no longer appear. This produces a target description and cached reasoning trace. The resulting seed ranking is important because its top-1 is often strong even when the later top-k is incomplete.

Before introducing the visual retrieval route, we apply a top-1 VLM slot selector inside the existing candidate list. This stage compares candidate slot evidence against the edit constraints and promotes a challenger only when its slot score is clearly better than the current top-1. It does not use DFN visual retrieval or add new visual candidates; its purpose is to correct obvious top-1 mistakes while keeping the original top-k recall fixed.

We also tested broader text reranking with multiple descriptions and edit-type-specific weighting. These variants changed many top-1 predictions and disturbed the top-50 ordering, but they did not improve the hidden-test score. We therefore use text reasoning mainly as a stable seed route, not as the final reranker.

\subsection{Route B: Visual Candidate Route}

The second route is visual-first. We sample each candidate video into a contact sheet and embed the sheets with DFN-H and DFN-L visual encoders. Query prompts are derived from the edit and target reasoning, then used to retrieve visually compatible candidates from the gallery. This route is designed to recover targets that are not ranked highly by the reasoning/text route.

The visual route is intentionally not trusted as a direct top-1 selector. Instead, its role is to add missing candidates into the top-10/top-5 region. This design raises high-rank recall while preserving the seed route's current top-1.

\subsection{Top-10 Merge}

The two routes are merged into a top-10 candidate set. The merge keeps the current top-1 fixed, retains high-confidence seed candidates, and injects visual candidates that are new relative to the seed top-k. Duplicates and the reference source video are removed before constructing the ranked output. This merged set is the main candidate pool for final reranking.

The key practical choice is to merge before selecting. Directly replacing the top-1 with a visual winner can reduce R@1, while preserving top-1 and injecting visual candidates strongly improves R@5/R@10/R@50. The final decision is therefore deferred to a separate reranking stage.

\subsection{1v1 VLM Final Reranker}

The final reranker compares the current top-1 against one challenger at a time. For each comparison, the VLM sees the reference video contact sheet, the edit instruction, target reasoning fields, the current top-1 contact sheet, and the challenger contact sheet. It is instructed to choose the challenger only if the challenger clearly satisfies the edit and the current top-1 has decisive visual errors.

The 1v1 format is central. A direct top-5 classification prompt is too easy to bias toward local visual similarity. In contrast, a pairwise prompt frames the decision as a replacement test: the challenger must beat an already strong current top-1, not merely look plausible.

\subsection{Risk-Controlled Promotion}

Accepted promotions must satisfy conservative checks: high confidence, low replacement risk, explicit fatal errors in the current top-1, and no fatal errors in the challenger. The full system adds a structured consensus filter in which a slot verifier and a fatal-error ranking reviewer must agree on the same non-top1 candidate. This final filter changes only a small number of top-1 predictions and leaves the top5/top10/top50 sets unchanged.

\section{Implementation and Experiments}

\subsection{Implementation Details}

The full system is zero-shot: no supervised training or fine-tuning is used. All VLM captioning, reasoning, reranking, and audit calls use {gemini-3.1-pro-preview} ~\cite{gemini}. Candidate and query videos are rendered as contact sheets with 8 uniformly sampled frames, 4 columns, and 320$\times$220 tiles. Candidate captioning uses temperature 0.1, 900 output tokens, and a 512-token thinking budget; query reasoning uses temperature 0.2, 1400 output tokens, and a 1600-token thinking budget.

The initial text seed is a local TF-IDF ranker over Gemini-generated candidate descriptions and query target descriptions. It combines word 1--2 grams and character 3--5 grams with weights 0.7 and 0.3, followed by L2 normalization and cosine scoring, and outputs 50 predictions per query. The top-1 slot selector uses Gemini at temperature 0.0 with 6000 output tokens and a 2048-token thinking budget; promotion requires a non-top1 challenger with no fatal missing/violation and at least a 5-point slot-score margin over the current top-1.

The visual route uses DFN-H and DFN-L image encoders (ViT-H-14-378-quickgelu/dfn5b and ViT-L-14-quickgelu/dfn2b), sourced from OpenCLIP ~\cite{ilharco_gabriel_2021_5143773} based on the Data Filtering Networks~\cite{fang2024data}. Image and text embedding batch sizes are 96 and 32. Each visual model retrieves top-100 candidates; the VLM tournament considers at most 18 candidates, including the top 14 visual candidates, the current top-1, current top-10, V1 top-15, and query-rewrite top-10. The final visual-injection submission preserves the current top-1 and injects visual candidates into the top-5/top-10 region.

The final 1v1 reranker again uses Gemini at temperature 0.0, with 5000 output tokens and a 1024-token thinking budget. It compares the current top-1 against new visual top2--top5 challengers; the hidden-test run contains 1,008 pairwise comparisons over 299 queries. Strict promotion requires high confidence, low replacement risk, score margin $>5$, at least one fatal error in the current top-1, and zero fatal errors in the challenger. The structured consensus audit uses top-5 candidates, temperature 0.0, 7000 output tokens, and a 2048-token thinking budget; it performs two independent judgments for each of the 301 hidden-test queries and accepts only same-winner, high-confidence, low-risk, zero-fatal challengers.

\subsection{Evaluation Protocol}

CoVR-R evaluates retrieval from a reference video and edit instruction to a target video. The system outputs 50 predicted target ids per query, and the evaluator reports R@1, R@5covr, R@10, and R@50. The hidden test split contains 301 queries after validation. We do not report full-pipeline validation reranking results because the Gemini-based pairwise and consensus stages require a large API budget when run over an additional split.

\subsection{Hidden Test Results}

Table~\ref{tab:main-results} reports the hidden-test results by main method module. The text seed uses Gemini-generated descriptions with text-based ranking. The top-1 selector then improves the seed ranking within the existing candidate list. Visual injection adds DFN-retrieved candidates into the high-rank list while preserving the current top-1. The final system keeps the stronger candidate set and applies conservative VLM consensus promotion for top-1 replacement.

\begin{table*}[t]
\centering
\caption{Hidden-split performance of the main method modules.}
\label{tab:main-results}
\small
\setlength{\tabcolsep}{4pt}
\begin{tabular}{p{0.2\linewidth}p{0.42\linewidth}cccc}
\toprule
Stage & Policy & R@1 & R@5 & R@10 & R@50 \\
\midrule
Text seed & Gemini-generated descriptions with text-based ranking. & 68.14 & 91.41 & 94.28 & 97.65 \\
Top-1 selector & VLM slot selection inside the existing candidate list. & 88.38 & 91.41 & 94.28 & 97.65 \\
Visual injection & DFN visual-route candidate injection while preserving the current top-1. & 91.41 & 97.14 & 97.81 & 98.82 \\
\textbf{Final system} & \textbf{Visual injection plus conservative VLM consensus promotion.} & \textbf{95.28} & \textbf{97.98} & \textbf{98.48} & \textbf{99.66} \\
\bottomrule
\end{tabular}
\end{table*}

\begin{figure}[t]
\centering
\includegraphics[width=\linewidth]{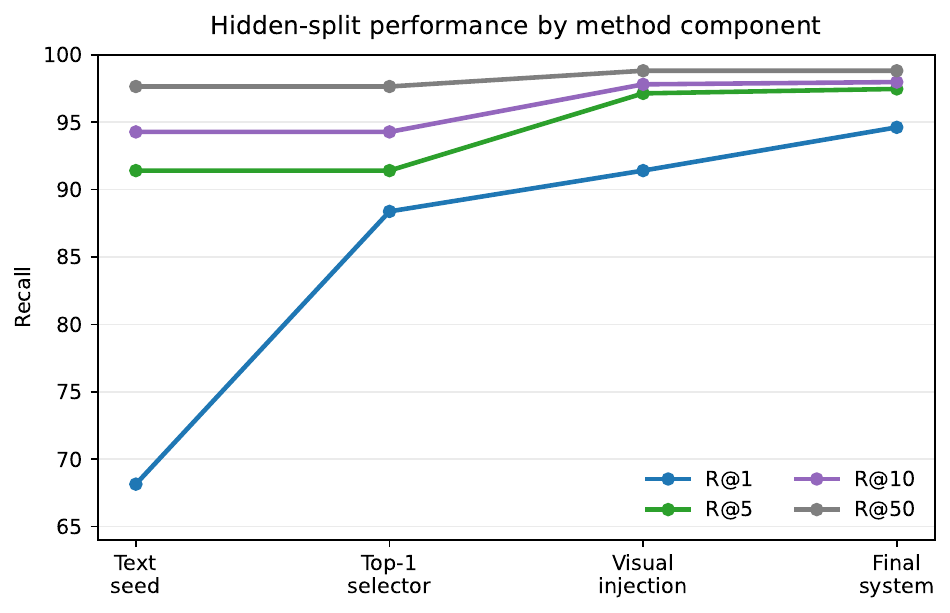}
\caption{Hidden-test progression across the main method modules.}
\label{fig:score-progression}
\end{figure}

The progression shows two different bottlenecks. Existing-candidate selection mainly improves R@1, visual injection mainly improves high-rank recall, and the final consensus stage improves R@1 on top of the stronger candidate set.

\subsection{Ablations and Observations}

The strongest pattern is that top-1 selection and high-rank recall improve in different stages. The non-DFN top-1 selector raises R@1 while R@5/R@10/R@50 remain fixed, showing pure selection errors inside the original list. Later, preserving the current top-1 while injecting visual candidates raises R@5 from 91.41 to 97.98 and R@50 from 97.65 to 99.66, while R@1 stays unchanged. Visual retrieval therefore recovers missing targets, but those targets still need a separate, safer selection mechanism.

\begin{figure}[t]
\centering
\includegraphics[width=\linewidth]{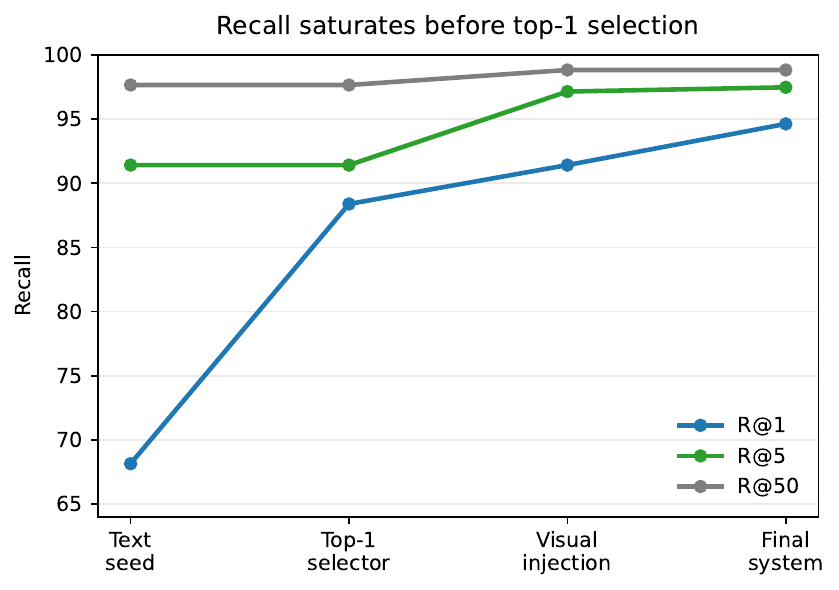}
\caption{R@5/R@50 saturate before R@1. The final gains therefore come from safer top-1 selection rather than simply retrieving more candidates.}
\label{fig:bottleneck}
\end{figure}

The 1v1 final reranker helps because it frames promotion as a strict replacement test: a challenger must satisfy the edit instruction and expose decisive errors in the current top-1. Direct multi-candidate VLM choice and text-only reranking were less stable, and aggressive promotion from top6--top50 reduced R@1. The final system therefore promotes only candidates already moved into the merged top-five/top-ten region and only when the consensus checks agree.

\section{Conclusion}

Our final CoVR-R system is a zero-shot staged retrieval and reranking pipeline. A reasoning/text route keeps a strong seed ranking, a VLM slot selector first corrects top-1 mistakes inside the existing candidate list, a visual route recovers missing candidates into the top-k set, and a 1v1 VLM final reranker performs conservative top-1 promotion. The final hidden-test result reaches 94.62 R@1 and 98.82 R@50. The central finding is that CoVR-R performance depends on separating candidate recall from top-1 replacement: improve the seed top-1 when the answer is already in the list, retrieve broadly with visual routes when recall is missing, then promote narrowly with pairwise evidence.

{
    \small
    \bibliographystyle{ieeenat_fullname}
    \bibliography{main}
}

\end{document}